\documentclass[11pt]{article}
\usepackage[final]{acl}
\usepackage{times}
\usepackage{latexsym}
\usepackage[T1]{fontenc}
\usepackage[utf8]{inputenc}
\usepackage{microtype}
\usepackage{inconsolata}
\usepackage{graphicx}
\usepackage{booktabs}
\usepackage{amsmath}
\usepackage{CJKutf8}
\usepackage{tabularx}
\usepackage{array}
\usepackage{xcolor}
\usepackage{multirow}
\usepackage{tcolorbox}
\usepackage{enumitem}
\usepackage{algorithm}
\usepackage{algpseudocode}

\tcbuselibrary{skins,breakable}
\newtcolorbox{promptbox}[2][]{
  enhanced,
  colback=blue!3,
  colframe=blue!45!black,
  coltitle=white,
  colbacktitle=blue!55!black,
  title={#2},
  fonttitle=\bfseries\small,
  fontupper=\footnotesize,
  sharp corners,
  boxrule=0.5pt,
  top=2pt, bottom=2pt,
  left=3pt, right=3pt,
  #1
}

\newcommand{\up}[1]{\rlap{\,{\scriptsize\textcolor{blue}{+#1}}}}
\newcommand{\dn}[1]{\rlap{\,{\scriptsize\textcolor{red}{-#1}}}}
\title{Quantifying and Mitigating Korean Jamo-Level Typographical Vulnerabilities in Large Language Models}

\author{
\textbf{Seojin Lee} \quad
\textbf{Hwanhee Lee}\thanks{\;\;Corresponding author.} \\
Chung-Ang University, Seoul, Korea \\
\texttt{\{seojin0311,hwanheelee\}@cau.ac.kr} \\
\url{https://github.com/SJLee0311/korean-jamo-typo}
}

\begin{document}
\maketitle

\begin{abstract}
Korean introduces an additional typographical perturbation level not captured by ordinary character-level edit models: because syllable blocks are internally composed of sub-character units called jamo, keyboard-level errors can occur \textit{within} a syllable, either producing a valid but semantically altered character or exposing raw jamo on the surface. 
Both outcomes disrupt sub-word tokenization and are not reliably corrected by existing grammatical error correction pipelines, leaving LLMs directly exposed to corrupted inputs. 
To quantify this vulnerability, we apply five jamo-level perturbation types to the KMMLU benchmark and evaluate four language models, finding that accuracy declines monotonically with perturbation intensity and that parameter scaling does not confer robustness against intra-syllabic noise. 
We further show that typo-corrupted inputs induce a distinct shift in internal representations that is not reducible to ordinary answer incorrectness, and that a simple linear probe trained on these representations detects unseen perturbation types with high AUROC. 
Motivated by this signal, we propose \textbf{Typo-Aware Chain-of-Thought (TACoT)}, which routes inputs to chain-of-thought inference only when the probe detects a likely typo, recovering a substantial portion of the CoT accuracy gain at a fraction of the inference cost.
\end{abstract}

\begin{figure}[t]
    \centering
    \includegraphics[width=\columnwidth]{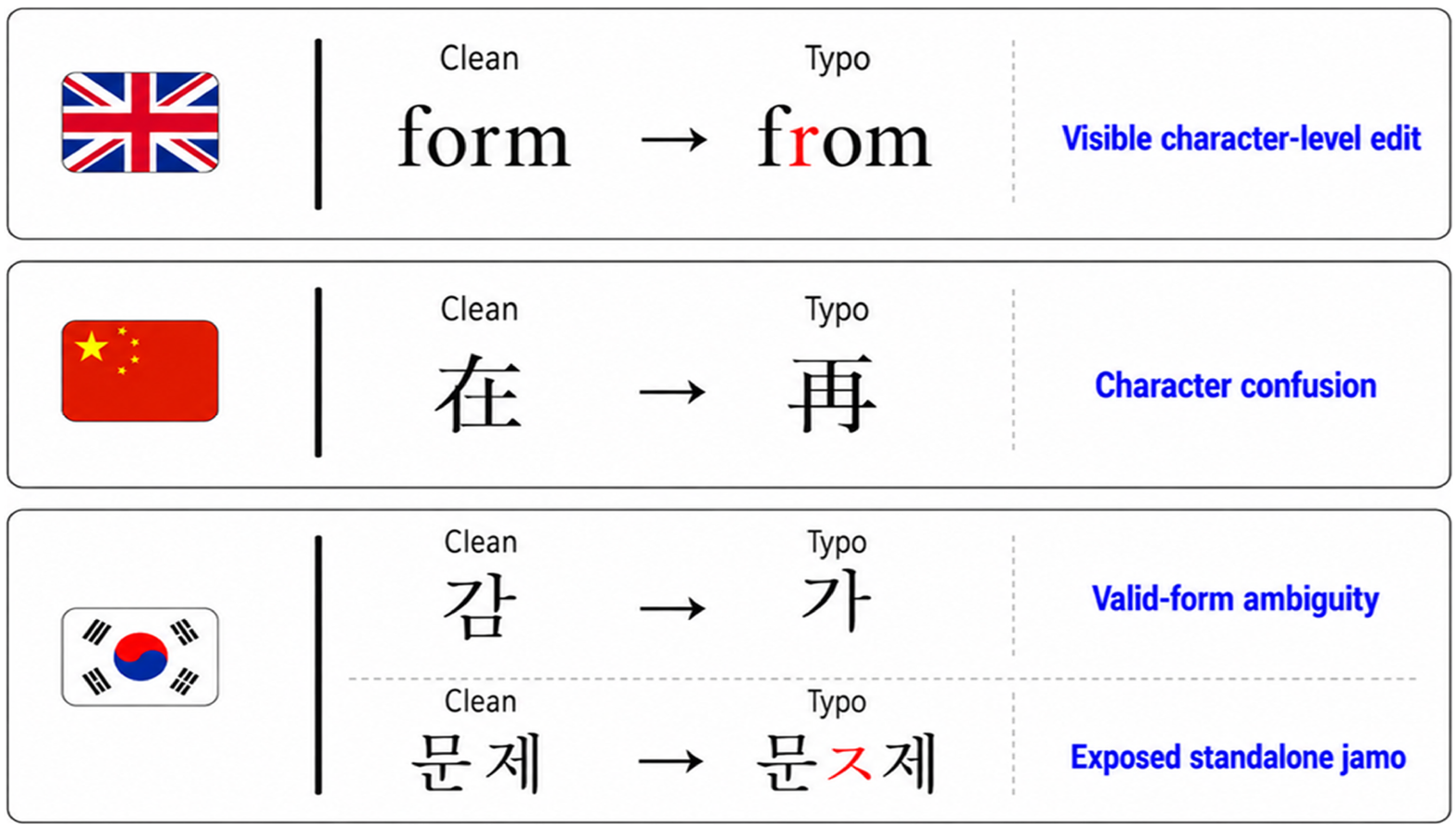}
    \vspace{-10mm}
    \caption{Typographical perturbations across writing systems. In English and Chinese, errors occur at the visible character level, whereas Korean errors occur within syllable blocks and surface as one of two failure modes: valid-form ambiguity or exposed standalone jamo that break the syllabic structure.}
    \label{fig:korean_typo_concept}
    \vspace{-3mm}
\end{figure}

\section{Introduction}
Large Language Models (LLMs) have achieved remarkable success across diverse natural language processing tasks, leading to their widespread deployment. 
However, their robustness against input noise remains a critical bottleneck. 
Among various user-generated noises, typographical errors are ubiquitous.
Characterizing and mitigating LLM brittleness under typographical perturbations has thus emerged as an essential line of trustworthy AI research~\citep{gao2018black,zhu2024promptrobust,zhu2024promptbench}. 
Nevertheless, existing robustness benchmarks are predominantly English-centric, modeling typos merely as surface-character edits~\citep{gao2018black,zhu2024promptrobust,zhu2024promptbench,gan2024reasoning}, a paradigm that fails to capture intra-syllabic perturbations unique to scripts like Korean.

While classical spelling correction targets linear edit distances~\citep{kukich1992techniques,brill-moore-2000-improved} and Chinese systems focus on phonologically or visually similar character substitutions~\citep{cheng-etal-2020-spellgcn,sun-etal-2021-chinesebert}, the Korean writing system introduces an additional perturbation level beyond visible character edits. 
Korean is a structured alphabetic syllabary where individual letters called \textit{jamo} are assembled into syllable blocks. 
Because keyboard input is typically produced through individual jamo keystrokes, realistic Korean typos occur at the intra-syllabic level rather than the surface syllable level~\citep{kwon2021handling}.

These intra-syllabic perturbations pose a critical challenge: the lack of an effective correction prior. 
As shown in Figure~\ref{fig:korean_typo_concept}, a jamo-level perturbation may convert a syllable into a contextually incorrect valid character (causing semantic ambiguity) or expose standalone jamo on the surface, breaking the syllabic structure. 
Both outcomes disrupt sub-word tokenizers optimized for well-formed syllabic blocks. 
Furthermore, grammatical error correction pipelines are not designed for intra-syllabic perturbations and may inadvertently alter sentence semantics~\citep{kim-etal-2024-kogec}. 
Consequently, LLMs must directly process these corrupted representations, rendering them vulnerable.

To systematically understand and combat this vulnerability, we investigate the impact of Korean typographical errors on LLMs by addressing two research questions: 
1) How vulnerable are state-of-the-art LLMs to realistic jamo-level perturbations? 
2) Do LLMs implicitly recognize such structural noise within their internal representations, and can we leverage these signals for efficient defense?

To answer these questions, we construct a Korean typo benchmark applying five keyboard-derived perturbation types to the KMMLU evaluation suite~\citep{son2025kmmlu}. 
Across four language models, accuracy declines monotonically with perturbation intensity, confirming that parameter scaling fails to alleviate intra-syllabic brittleness. 
Moreover, our probing analysis reveals that typo-corrupted inputs induce a distinct representational shift that is not reducible to ordinary answer incorrectness, and that a simple linear probe detects this signal across held-out error types with high AUROC. 
Motivated by this signal, we propose \textbf{Typo-Aware Chain-of-Thought (TACoT)}, which employs the lightweight internal probe to monitor representations at inference time, invoking CoT only when a typographical anomaly is detected. 
Our results show that this typo-aware routing substantially recovers CoT accuracy while reducing inference cost on clean inputs.

In summary, our contributions are as follows:
\begin{itemize}[leftmargin=*, itemsep=2pt, topsep=2pt]
    \item We formalize a five-type jamo-level perturbation framework based on Korean keyboard mechanics, capturing structural anomalies that lack standard correction priors.
    \item We establish a controlled benchmark to quantify accuracy drops across diverse LLMs, showing that parameter scaling alone fails to resolve intra-syllabic brittleness.
    \item We discover a distinct typo signal in LLM internal representations and introduce a Typo-Aware CoT framework that leverages this signal for efficient, low-cost routing defense.
\end{itemize}

\section{Related Work}
\label{sec:related_work}

\subsection{Typographical Robustness in LLMs}
Character-level perturbations have long been used to study the brittleness of neural NLP models. 
DeepWordBug showed that small black-box character edits can substantially degrade text classifiers \citep{gao2018black}. 
With the rise of LLMs, PromptRobust and PromptBench extended robustness evaluation to prompt perturbations, including typo-like changes \citep{zhu2024promptrobust, zhu2024promptbench}.
More recent work studies typographical errors in LLM-specific settings: \citet{gan2024reasoning} show that adversarial typos can disrupt CoT reasoning trajectories, while \citet{zhao2026evaluating} evaluate LLM robustness against multilingual keyboard-based typographical errors. 
These studies establish typo robustness as an important failure mode for LLMs, but they primarily model typos as visible character-level perturbations. 
Our work focuses on Korean, where typographical errors can also arise within the internal composition of syllable blocks.

\subsection{Korean Grammatical Error Correction}
Korean grammatical error correction (GEC) aims to detect and correct erroneous Korean text, including grammatical, spacing, punctuation, and spelling errors. 
Prior work has developed Korean GEC datasets, annotation schemes, and error taxonomies \citep{yoon-etal-2023-towards}, as well as neural correction models based on Transformer architectures, copying mechanisms, and synthetic grammatical noise \citep{lee2021korean}. KoGEC further formulates Korean GEC as a translation-style correction problem by fine-tuning pre-trained multilingual translation models \citep{kim-etal-2024-kogec}. 
These studies provide important resources and models for Korean text correction. 
However, their primary goal is to restore well-formed text, whereas our work studies how controlled Korean typos affect downstream LLM behavior. In particular, we focus on jamo-level perturbations that alter the internal composition of Korean syllable blocks, quantify their impact on LLM accuracy, and analyze whether they induce detectable hidden-state signals.

\tcbset{
  repairedstyle/.style={colback=green!3, colframe=green!45!black, colbacktitle=green!45!black},
  unchangedstyle/.style={colback=orange!4, colframe=orange!65!black, colbacktitle=orange!70!black},
  lossystyle/.style={colback=red!3, colframe=red!55!black, colbacktitle=red!60!black}
}
\begin{figure*}[!ht]
\centering
\begin{CJK}{UTF8}{mj}
\begin{minipage}[t]{0.31\textwidth}
\begin{promptbox}[repairedstyle]{Repaired}
\textbf{Clean}\\[2pt]
유형자산의 원가를 구성하는 것은?\\
\textit{(What constitutes the cost of tangible assets?)}\\[4pt]
\textbf{Typo input}\\[2pt]
유형자산의 원가를 구\textcolor{red}{ㅅ}성하는 것\textcolor{red}{ㅇ}은?\\[4pt]
\textbf{KoGEC output}\\[2pt]
유형자산의 원가를 구성하는 것은?
\end{promptbox}
\end{minipage}
\hspace{0.01\textwidth}
\begin{minipage}[t]{0.31\textwidth}
\begin{promptbox}[unchangedstyle]{Left unchanged}
\textbf{Clean}\\[2pt]
주식배당에 관한 설명 중 옳지 않은 것은?\\
\textit{(Which description of stock dividends is incorrect?)}\\[4pt]
\textbf{Typo input}\\[2pt]
주\textcolor{red}{시}배당에 관한 \textcolor{red}{서}명 중 \textcolor{red}{오}지 \textcolor{red}{아}은 것은?\\[4pt]
\textbf{KoGEC output}\\[2pt]
주시배당에 관한 서명 중 오지 아은 것은?
\end{promptbox}
\end{minipage}
\hspace{0.01\textwidth}
\begin{minipage}[t]{0.31\textwidth}
\begin{promptbox}[lossystyle]{Lossy rewrite}
\textbf{Clean}\\[2pt]
회계정보와 관련한 설명으로 옳지 않은 것은?\\
\textit{(Which description of accounting information is incorrect?)}\\[4pt]
\textbf{Typo input}\\[2pt]
회\textcolor{red}{ㅖㄱ}정보와 관련한 설명으\textcolor{red}{ㅗㄹ} 옳\textcolor{red}{ㅣㅈ} 않\textcolor{red}{ㅇㄴㅡ} 것은?\\[4pt]
\textbf{KoGEC output}\\[2pt]
정보와 관련한 설명은 옳지 않은 것은?
\end{promptbox}
\end{minipage}
\end{CJK}
\caption{Representative outputs from the KoGEC model on typo-perturbed KMMLU questions. The model can repair simple exposed-jamo errors (Repaired), but may leave valid-form typos unchanged (Left unchanged) or rewrite heavily corrupted inputs into fluent yet semantically altered text (Lossy rewrite).}
\label{fig:gec_examples}
\end{figure*}

\section{Korean Typographical Errors}
\label{sec:korean_typographical_errors}

\subsection{Intra-Syllabic vs. Character-Level Errors}
\label{sec:typos_across_writing_systems}

Conventional spelling correction models typos as visible character edits under edit distance assumptions~\citep{kukich1992techniques,brill-moore-2000-improved}, and Chinese spelling correction targets phonologically or visually similar character substitutions using pinyin, glyph, or radical information~\citep{cheng-etal-2020-spellgcn,sun-etal-2021-chinesebert}.
In both cases, the error unit aligns with a visible character, giving correction systems a stable target for recovery.

Korean shares this character-level typo space, but also introduces an intra-syllabic perturbation level. 
Since Korean syllable blocks are composed of jamo (onset, nucleus, and optional coda) and keyboard input is produced through individual jamo keystrokes, typographical errors can occur \textit{within} a syllable block.
A jamo-level perturbation may yield another well-formed Korean syllable (e.g., \begin{CJK}{UTF8}{mj}감 $\rightarrow$ 가\end{CJK}), producing valid-form ambiguity, or may break the syllable block and expose standalone jamo on the surface (e.g., \begin{CJK}{UTF8}{mj}문제 $\rightarrow$ 문ㅈ제\end{CJK}), producing a structurally degraded sequence with no stable character-level recovery unit~\citep{kwon2021handling}. 
Figure~\ref{fig:korean_typo_concept} illustrates this distinction; the resulting orthographic disruption is invisible to standard character-level correction pipelines and to tokenizers optimized for well-formed syllabic blocks.

\subsection{Limitations of Korean GEC}
\label{sec:correction_limits}

Whether typo-corrupted inputs can be automatically corrected prior to LLM inference remains an open question. 
However, our typographical setting fundamentally differs from standard GEC. 
While conventional GEC systems are designed to rectify grammaticality, spacing, punctuation, and surface spelling, our perturbations specifically target the internal jamo composition of Hangul syllables. 
Consequently, valid-form perturbations yield another well-formed Korean surface, making the error difficult to flag as ungrammatical. 
Conversely, exposed-jamo sequences break the syllable structure altogether, producing degraded strings that do not correspond cleanly to the word- or syllable-level edits assumed by standard correction systems.

To analyze this mismatch, we evaluate KoGEC~\citep{kim-etal-2024-kogec} on typo-perturbed KMMLU questions in Figure~\ref{fig:gec_examples}.
While KoGEC repairs simple exposed-jamo errors, it leaves valid-form perturbations unchanged or rewrites heavily corrupted inputs into semantically altered text.
For instance, severe corruptions lead to the loss of the semantic subject, shifting the question's core meaning.
Consequently, Korean jamo typos are not reliably reducible to existing GEC pipelines, demonstrating that pre-correction does not provide a stable solution to this robustness bottleneck.

\section{Korean Typographical Perturbation Framework}

\subsection{Evaluation Benchmark}

We use KMMLU~\citep{son2025kmmlu} as the main evaluation benchmark.
KMMLU consists of 35,030 expert-level four-choice questions collected from original Korean examinations.
We apply typo perturbations only to the question; answer choices are kept unchanged to ensure that accuracy loss reflects input comprehension failure rather than output matching artifacts.
Further dataset details are provided in Appendix~\ref{sec:appendix_datasets}.

\subsection{Typo Taxonomy}
 
We define five typo types grounded in realistic keyboard input errors on a standard Korean layout, each targeting a different aspect of the jamo--syllable structure.
Table~\ref{tab:taxonomy} summarizes the types with examples derived from the sentence \textit{``\begin{CJK}{UTF8}{mj}감기가 다 낫다니 다행이야\end{CJK}''} (``It's great that you've recovered from your cold.'').
 
\begin{table}[h]
\centering
\small
\begin{tabularx}{\columnwidth}{lX}
\toprule
\textbf{Typo Type} & \textbf{Perturbed} \\
\midrule
Jamo Substitution  & \begin{CJK}{UTF8}{mj}\textcolor{red}{검}기가 다 낫다니 다행이야\end{CJK} \\
Jongseong Deletion & \begin{CJK}{UTF8}{mj}\textcolor{red}{가}기가 다 낫다니 다행이야\end{CJK} \\
Jamo Repetition    & \begin{CJK}{UTF8}{mj}\textcolor{red}{ㄱ}감기가 다 낫다니 다행이야\end{CJK} \\
Space Deletion     & \begin{CJK}{UTF8}{mj}감기가\textcolor{red}{다}낫다니 다행이야\end{CJK} \\
Jamo Transposition & \begin{CJK}{UTF8}{mj}\textcolor{red}{ㄱㅁㅏ}기가 다 낫다니 다행이야\end{CJK} \\
\midrule
\multicolumn{2}{l}{Clean: \begin{CJK}{UTF8}{mj}감기가 다 낫다니 다행이야\end{CJK}} \\
\bottomrule
\end{tabularx}
\caption{Korean typographical error taxonomy. Perturbed characters are shown in {\color{red}red}. Original sentence shown at the bottom.}
\label{tab:taxonomy}
\end{table}
 
\paragraph{Jamo Substitution.}
A jamo is replaced by one from an adjacent key, frequently yielding another valid Korean syllable and making error detection ambiguous (e.g., \begin{CJK}{UTF8}{mj}감\end{CJK} $\rightarrow$ \begin{CJK}{UTF8}{mj}검\end{CJK}).

\paragraph{Jongseong Deletion.}
The jongseong of a syllable is dropped (e.g., \begin{CJK}{UTF8}{mj}값 $\rightarrow$ 가\end{CJK}), always yielding a well-formed Korean syllable and thus being particularly difficult to detect.

\paragraph{Jamo Repetition.}
A jamo component is duplicated, surfacing as a raw jamo adjacent to the host syllable. Such sequences fall outside standard syllable blocks and are invisible to word-level spell checkers.

\paragraph{Space Deletion.}
Inter-eojeol spaces are removed, concatenating adjacent spacing units (eojeol are Korean orthographic spacing units, roughly analogous to words). 
Unlike the other types, this does not alter the internal jamo composition of syllables.

\paragraph{Jamo Transposition.}
The order of jamo within a syllable is swapped (e.g., \begin{CJK}{UTF8}{mj}감 $\rightarrow$ ㄱㅁㅏ\end{CJK}), dissolving the onset--nucleus--coda structure and exposing raw jamo poorly aligned with standard vocabularies.

This taxonomy is not an arbitrary construction: it covers the large majority of real Korean input errors observed in practice, as we verify in Appendix~\ref{sec:appendix_taxonomy_validation}.

\subsection{Perturbation Intensity}
 
For each typo type, we define five intensity levels corresponding to perturbation rates of $5\%$ through $25\%$ of the Korean syllables in the question field.
Each level is applied independently from the original clean input, so types are never mixed within a single instance.\footnote{For Space Deletion, intensity is measured over the number of spaces rather than Korean syllables.}
Within each level, the affected syllables are chosen at random. 
We confirm in Appendix~\ref{sec:appendix_robustness} that this randomness does not affect our conclusions.
This yields $35{,}030 \times 5 \times 5 = 875{,}750$ perturbed instances in total.

\begin{figure*}[!htbp]
    \vspace{-8pt}
    \centering
    \includegraphics[width=0.48\linewidth]{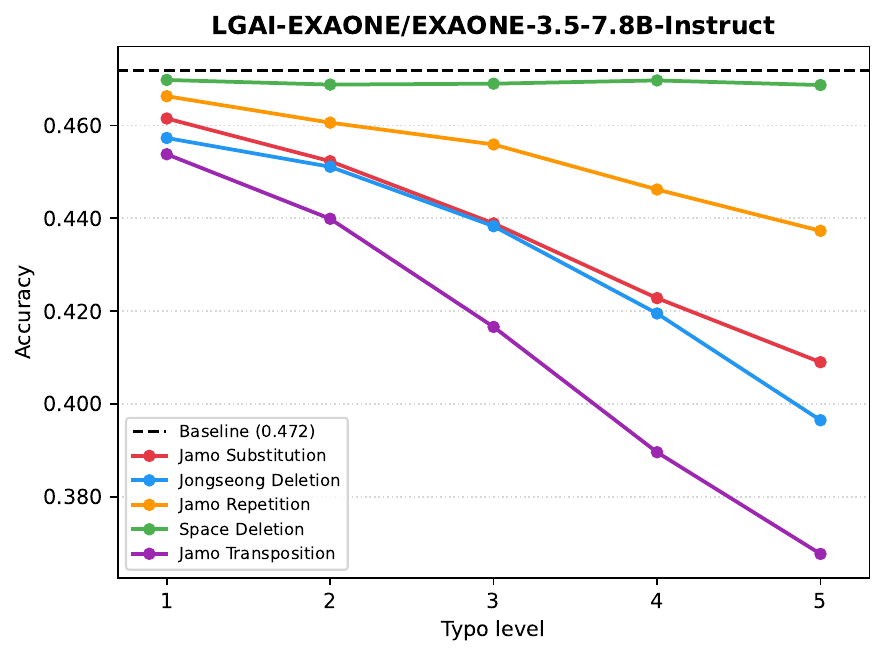}\hfill
    \includegraphics[width=0.48\linewidth]{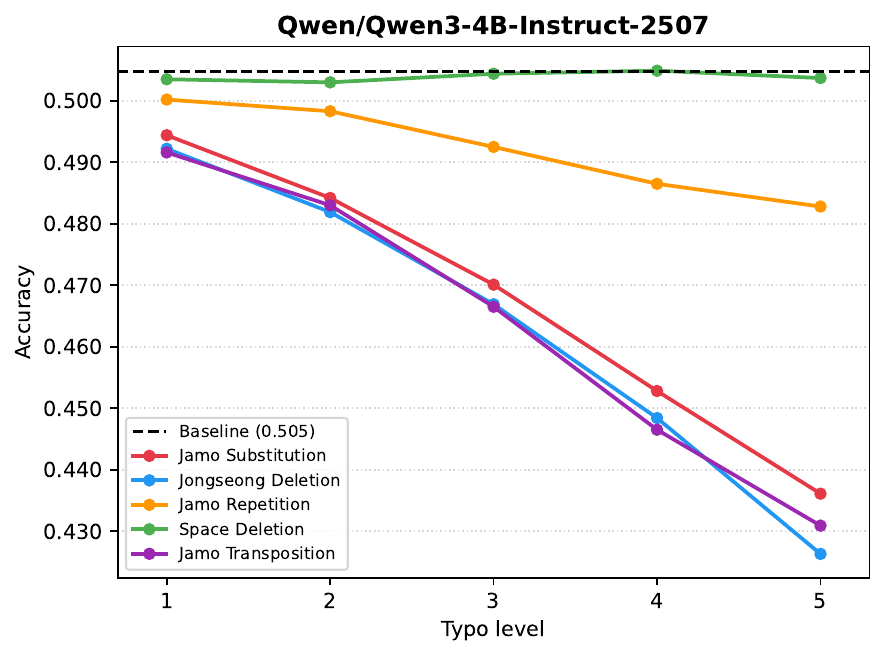}\\[6pt]
    \includegraphics[width=0.48\linewidth]{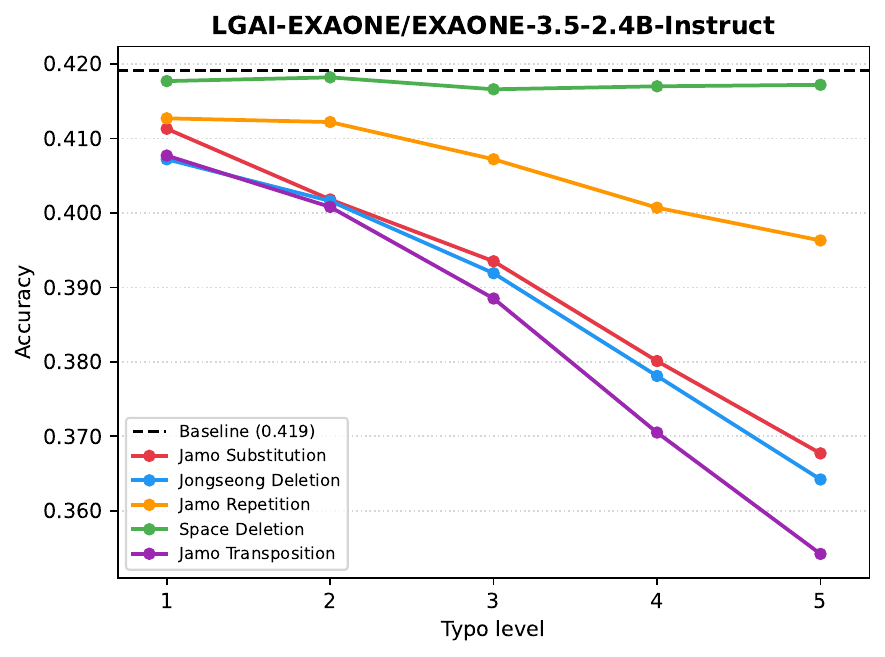}\hfill
    \includegraphics[width=0.48\linewidth]{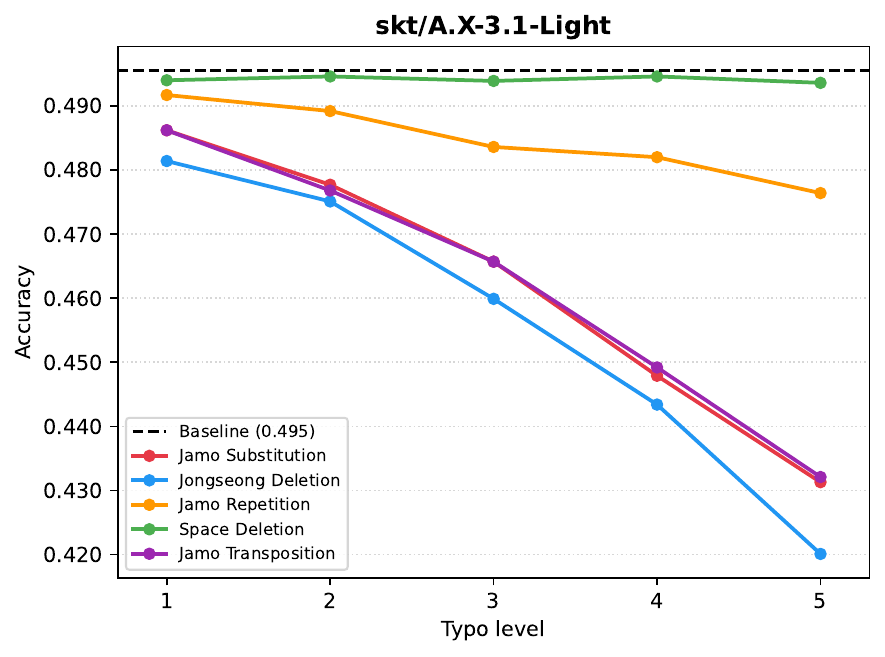}
    \vspace{-4pt}
    \caption{KMMLU accuracy across five typo intensity levels for all four models. Dashed lines indicate clean baseline accuracy.}
    \label{fig:results}
    \vspace{-8pt}
\end{figure*}

\section{Impact of Korean Typographical Errors}
\label{sec:impact}
\subsection{Experimental Setup}
\paragraph{Models.}
We evaluate four models spanning a range of sizes and Korean language orientations.
EXAONE-3.5-2.4B-Instruct and EXAONE-3.5-7.8B-Instruct are bilingual (Korean--English) instruction-tuned models developed by LG AI Research, designed for real-world instruction following with strong Korean proficiency \citep{an2026exaone35serieslarge}.
A.X-3.1-Light is a Korean sovereign AI model developed entirely from scratch by SK Telecom, trained on a Korean-focused multilingual corpus.\footnote{https://huggingface.co/skt/A.X-3.1-Light}
Qwen3-4B-Instruct-2507 is a general-purpose multilingual model by Alibaba; we use the 2507 checkpoint, which features significant improvements in instruction following and multilingual comprehension \citep{yang2025qwen3}.
This model serves as a non-Korean-specialized baseline to probe whether typo robustness is tied to Korean-specific training.
 
\paragraph{Implementation Details.}
All models are evaluated on KMMLU in a zero-shot setting.
We use greedy decoding with temperature 0 and \texttt{max\_new\_tokens} set to 8; unless otherwise noted, all experiments follow this configuration. Full prompt and inference details are provided in Appendix~\ref{sec:appendix_prompt}.

\subsection{Results}
Figure~\ref{fig:results} shows accuracy across five typo intensity levels for all four open-source models.
Across all four models, accuracy declines monotonically as intensity increases, confirming that Korean typographical errors consistently degrade performance regardless of model size or Korean specialization.
Jamo-level perturbations induce the largest drops: at the highest intensity, accuracy falls by up to 10.0\%p for EXAONE-7.8B and 7.0\%p for Qwen3-4B relative to their clean baselines.
Notably, model size does not correlate with robustness: EXAONE-7.8B exhibits a larger drop than its 2.4B counterpart under Jamo Transposition, suggesting that scale alone does not confer immunity to intra-syllabic perturbations.

Space Deletion stands out as a consistent exception: across all models, accuracy under space deletion remains close to the clean baseline even at the highest intensity.
While perceptually disruptive to human readers, Korean subword tokenizers appear to partially recover eojeol boundaries during segmentation, effectively absorbing the perturbation before it reaches the model's internal representations.
This divergence between human perception and model behavior highlights that the impact of a typo type cannot be inferred from surface-level intuition alone.

These results establish that even well-performing LLMs are meaningfully vulnerable to naturally occurring Korean typographical errors. Critically, this vulnerability persists regardless of model scale or Korean specialization, suggesting that the problem is structural rather than a matter of insufficient training. We therefore turn to the models' internal representations to ask whether this structural disruption leaves a detectable signal. We further evaluate larger-scale models in Appendix~\ref{sec:appendix_robustness}.

\section{Probing Internal Representations of Korean Typos}
\label{sec:hidden_signals}

The previous section shows that Korean typographical errors consistently degrade LLM accuracy. We now examine whether this degradation is reflected in the models' internal representations. If typos merely push the model toward an ordinary incorrect-answer state, then hidden states under typo inputs should largely align with clean inputs that the model already answers incorrectly. In contrast, if typo inputs induce a distinct internal state, we should observe a separable hidden-state shift that is not reducible to answer correctness. If such a signal exists and is separable from ordinary answer failure, it could serve as a lightweight, training-free basis for targeted mitigation.

\subsection{Locating Typo-Sensitive Layers}

We first identify where typo-induced changes are most visible inside the model. 
For each model, we extract the last-token hidden state from every transformer layer on the General Knowledge subset of HAERAE Bench (HAERAE-GK), using clean inputs and their typo-perturbed counterparts (see Appendix~\ref{sec:appendix_datasets} for dataset details). 
To select a typo-sensitive layer without training a classifier, we compute a Fisher-style separation score~\citep{fisher1936use} for each layer:
\begin{equation}
\small
J(l)=
\frac{\|\mu_{\mathrm{typo}}^{(l)}-\mu_{\mathrm{clean}}^{(l)}\|_2^2}
{\mathrm{tr}(\Sigma_{\mathrm{clean}}^{(l)})+\mathrm{tr}(\Sigma_{\mathrm{typo}}^{(l)})}.
\label{eq:fisher}
\end{equation}
where $\mu_{\mathrm{clean}}^{(l)}$ and $\mu_{\mathrm{typo}}^{(l)}$ 
are the mean hidden states of clean and typo inputs at layer $l$, 
and the denominator measures within-class dispersion. Intuitively, 
this criterion favors layers where clean and typo representations 
are far apart while remaining internally compact. We use the 
Fisher-selected layer for each model in the remaining hidden-state 
analyses; layer-wise score curves and selected layer indices are 
provided in Appendix~\ref{sec:appendix_hidden}.

\begin{figure}[!ht]
\centering
\includegraphics[width=\columnwidth]{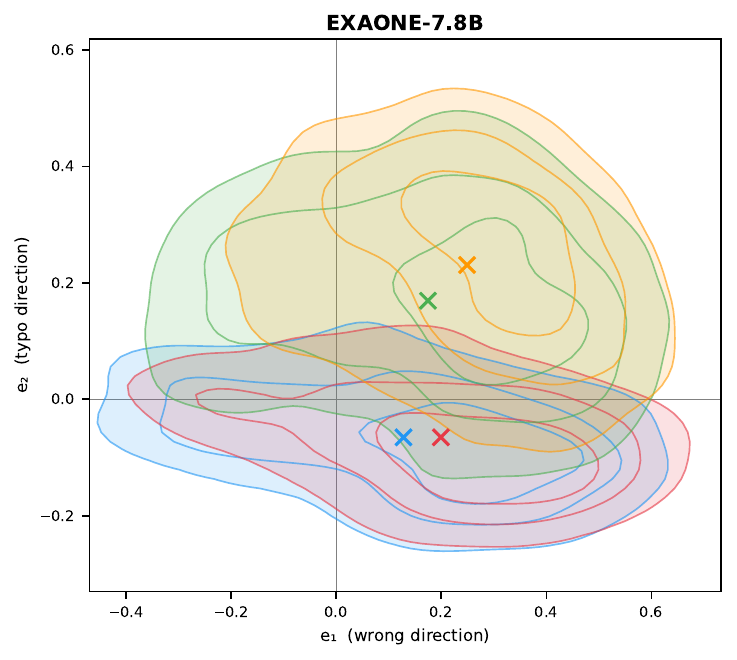}\\[4pt]
\includegraphics[width=\columnwidth]{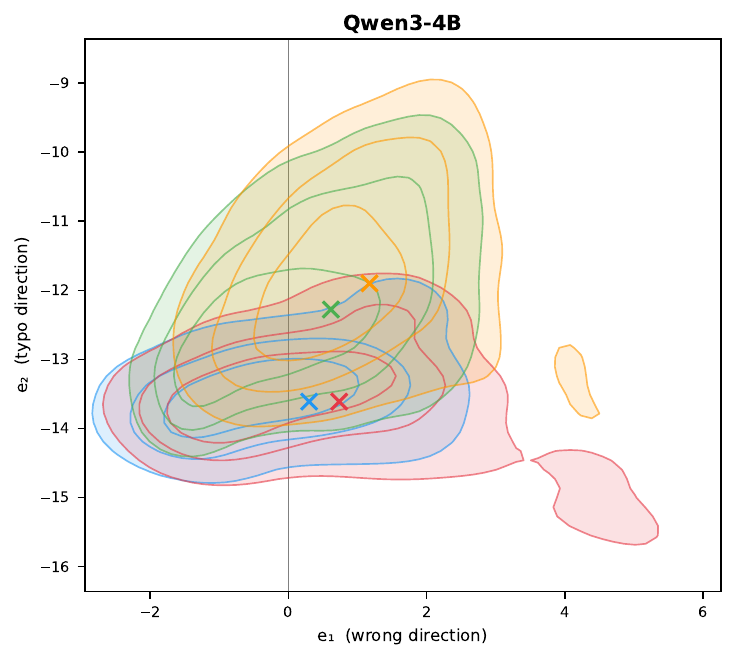}

\vspace{2pt}
\small
\textcolor{blue}{\rule{1em}{0.8pt}} Base-Correct \quad
\textcolor{red}{\rule{1em}{0.8pt}} Base-Wrong \quad \\
\textcolor{green!60!black}{\rule{1em}{0.8pt}} Typo-Correct \quad
\textcolor{orange}{\rule{1em}{0.8pt}} Typo-Wrong
\caption{Direction projection of hidden states at the Fisher-selected layer. Full projections for all models are provided in Appendix~\ref{sec:appendix_hidden}.}
\label{fig:direction_projection}
\end{figure}

\subsection{Typo Shift Is Distinct from Answer Incorrectness}

A typo input is typically both corrupted and answered incorrectly, so a possible concern is that the representational shift we observe may simply reflect answer failure: typo inputs may look different only because they cause the model to move toward the same representation as clean questions it answers incorrectly.
To test this, we compare two directions in the representation space at the Fisher-selected layer.
Let $\mu_{\mathrm{correct}}$ and $\mu_{\mathrm{wrong}}$ denote the mean hidden states of clean inputs the model answers correctly and incorrectly; we define the clean incorrectness direction as $d_{\mathrm{wrong}} = \mu_{\mathrm{wrong}} - \mu_{\mathrm{correct}}$ and the typo direction as $d_{\mathrm{typo}} = \mu_{\mathrm{typo}} - \mu_{\mathrm{correct}}$.
For visualization, we project hidden states onto a two-dimensional basis whose first axis is $d_{\mathrm{wrong}}$ and whose second axis is the component of $d_{\mathrm{typo}}$ orthogonal to $d_{\mathrm{wrong}}$.

Figure~\ref{fig:direction_projection} shows the resulting projections for clean-correct, clean-wrong, typo-correct, and typo-wrong examples.
Across models, typo examples shift along the typo direction regardless of whether the final answer is correct or wrong.
This indicates that typo inputs do not merely reproduce the hidden-state pattern of ordinary incorrect answers.
Instead, they induce a distinct representational displacement that coexists with, but is not reducible to, answer correctness.

\subsection{Hidden-State Typo Detection}
\label{sec:hs_detection}

The directional analysis suggests that Korean typo inputs leave a detectable representational signal.
We next test whether this signal can be used to identify typo inputs with a simple linear probe. 
For each model, we train a logistic regression classifier on the last-token hidden state at the Fisher-selected layer, using clean inputs and typo-perturbed inputs as binary labels, focusing on the four jamo-level typo types.\footnote{Space Deletion is excluded as it induces substantially weaker representational separation than jamo-level types; per-type Fisher scores are reported in Appendix~\ref{sec:appendix_hidden}.}

\begin{table}[tbh]
\centering
\small
\setlength{\tabcolsep}{3.5pt}
\begin{tabular}{lccccc}
\toprule
\textbf{Model} & \textbf{Sub.} & \textbf{Jong.} & \textbf{Rep.}
               & \textbf{Trans.} & \textbf{Avg.} \\
\midrule
EXAONE-2.4B & 0.912 & 0.885 & 0.927 & 0.971 & 0.924 \\
EXAONE-7.8B & 0.929 & 0.910 & 0.933 & 0.974 & 0.937 \\
A.X-Light   & 0.928 & 0.922 & 0.936 & 0.987 & 0.943 \\
Qwen3-4B    & 0.849 & 0.843 & 0.939 & 0.990 & 0.905 \\
\bottomrule
\end{tabular}
\caption{Held-out typo type detection AUROC using a logistic regression probe on Fisher-selected hidden states, pooled over all five perturbation intensities. Per-intensity results are reported in Table~\ref{tab:auroc_by_intensity}.}
\label{tab:hs_typo_detection}
\vspace{-3mm}
\end{table}

To test whether the probe captures a general typo signal rather than memorizing specific perturbation patterns, we use a held-out typo type evaluation. 
For each held-out jamo-level type, the classifier is trained on the remaining three types from HAERAE-GK and evaluated on the excluded type in KMMLU. 
Within each fold, we balance the training set by randomly sampling typo examples to match the number of clean inputs, drawing only from the non-held-out typo types. 
Thus, the test set differs both in questions and in typo type, and the held-out type is never used for probe training.

Table~\ref{tab:hs_typo_detection} reports AUROC on the held-out type. 
The probe achieves strong detection performance across all models, with mean AUROC ranging from 0.905 to 0.943. 
Jamo Transposition is consistently easiest to detect, while Jongseong Deletion is the hardest, mirroring their relative Fisher separation scores. 
These results show that typo-induced hidden-state shifts are not merely type-specific artifacts, but generalize to unseen jamo-level perturbation patterns.

\section{Mitigating Korean Typos via Internal Representations}
\label{sec:mitigation}

The previous section shows that Korean typos induce a detectable representational signal, especially for jamo-level perturbations that alter the internal composition of Korean syllables. 
We now use this signal for mitigation. 
A simple way to improve robustness is to ask the model to reason more carefully under typo-corrupted inputs. 
However, applying chain-of-thought (CoT) prompting~\citep{wei2022chain} to every input substantially increases output length and inference cost, including for clean or weakly affected inputs. 
We therefore introduce TACoT (Typo-Aware Chain-of-Thought), a probe-guided inference strategy that applies standard decoding by default and invokes CoT only when the hidden-state probe signals a likely typo.

\subsection{TACoT: Typo-Aware CoT}
\label{sec:typo-aware_cot}

Our typo-aware method uses the Fisher-selected layer identified in Section~\ref{sec:hidden_signals}. 
For each model, we train a logistic regression probe on the last-token hidden state from HAERAE-GK, using clean inputs and their typo-perturbed counterparts. 
For the typo class, we draw an equal number of examples from the pool of jamo-level perturbations, covering four typo types and five perturbation intensities, to prevent the probe from being dominated by the larger typo pool. 
The routing threshold is selected by maximizing Youden's $J$ statistic~\citep{youden1950index} on a held-out validation split; full training details are provided in Appendix~\ref{sec:appendix_probe}.

At inference time, each KMMLU input is first passed through the model to extract the last-token hidden state at the Fisher-selected layer. 
The probe then estimates $P(\mathrm{typo})$.
If this probability exceeds the learned threshold, we route the instance to CoT inference; otherwise, we use standard inference. 
Formally, the routing rule is
\[
\mathrm{route}(x)=
\begin{cases}
\mathrm{CoT}, & P(\mathrm{typo}\mid h_l(x)) \ge \theta, \\
\mathrm{Standard}, & P(\mathrm{typo}\mid h_l(x)) < \theta,
\end{cases}
\]
where $h_l(x)$ is the hidden state at the Fisher-selected layer and 
$\theta$ is the selected threshold.

\begin{table*}[tb]
\centering
\small
\setlength{\tabcolsep}{12pt}
\begin{tabular}{llcccccc}
\toprule
\textbf{Model} & \textbf{Method} & \textbf{Base} & \textbf{Sub.} & \textbf{Jong.} & \textbf{Rep.} & \textbf{Space.} & \textbf{Trans.} \\
\midrule
\multirow{5}{*}{EXAONE-2.4B}
  & Standard & 41.9 & 39.1 & 38.9 & 40.6 & 41.7 & 38.4 \\
  & Meta-Cognition & 42.1\up{0.2} & 39.2\up{0.1} & 38.9 & 40.7\up{0.1} & 41.9\up{0.2} & 38.6\up{0.2} \\
  & GEC & 41.9 & 38.5\dn{0.6} & 38.7\dn{0.2} & 40.7\up{0.1} & 41.5\dn{0.2} & 37.7\dn{0.7} \\
  & CoT & 45.2\up{3.3} & 41.9\up{2.8} & 41.5\up{2.6} & 43.9\up{3.3} & 44.9\up{3.2} & 41.1\up{2.7} \\
  & TACoT & 42.6\up{0.7} & 41.3\up{2.2} & 40.6\up{1.7} & 43.5\up{2.9} & 43.1\up{1.4} & 41.0\up{2.6} \\
\midrule
\multirow{5}{*}{EXAONE-7.8B}
  & Standard & 47.2 & 43.7 & 43.3 & 45.3 & 46.9 & 41.4 \\
  & Meta-Cognition & 47.4\up{0.2} & 43.5\dn{0.2} & 43.1\dn{0.2} & 44.9\dn{0.4} & 47.1\up{0.2} & 40.9\dn{0.5} \\
  & GEC & 47.2 & 42.7\dn{1.0} & 42.5\dn{0.8} & 45.6\up{0.3} & 46.9 & 41.9\up{0.5} \\
  & CoT & 49.7\up{2.5} & 46.9\up{3.2} & 46.2\up{2.9} & 48.6\up{3.3} & 49.4\up{2.5} & 44.8\up{3.4} \\
  & TACoT & 47.2 & 46.3\up{2.6} & 45.5\up{2.2} & 48.3\up{3.0} & 47.5\up{0.6} & 44.7\up{3.3} \\
\midrule
\multirow{5}{*}{A.X-Light}
  & Standard & 49.5 & 46.2 & 45.6 & 48.5 & 49.4 & 46.2 \\
  & Meta-Cognition & 50.8\up{1.3} & 46.9\up{0.7} & 46.0\up{0.4} & 49.3\up{0.8} & 50.6\up{1.2} & 47.0\up{0.8} \\
  & GEC & 49.5 & 45.0\dn{1.2} & 45.0\dn{0.6} & 47.9\dn{0.6} & 49.1\dn{0.3} & 44.0\dn{2.2} \\
  & CoT & 57.1\up{7.6} & 52.5\up{6.3} & 51.7\up{6.1} & 54.9\up{6.4} & 56.7\up{7.3} & 52.1\up{5.9} \\
  & TACoT & 49.6\up{0.1} & 49.8\up{3.6} & 48.9\up{3.3} & 52.8\up{4.3} & 49.7\up{0.3} & 51.0\up{4.8} \\
\midrule
\multirow{5}{*}{Qwen3-4B}
  & Standard & 50.5 & 46.8 & 46.3 & 49.2 & 50.4 & 46.4 \\
  & Meta-Cognition & 50.0\dn{0.5} & 45.8\dn{1.0} & 45.5\dn{0.8} & 48.4\dn{0.8} & 49.9\dn{0.5} & 45.3\dn{1.1} \\
  & GEC & 50.4\dn{0.1} & 46.3\dn{0.5} & 46.5\up{0.2} & 48.9\dn{0.3} & 50.0\dn{0.4} & 45.7\dn{0.7} \\
  & CoT & 53.7\up{3.2} & 47.9\up{1.1} & 47.5\up{1.2} & 51.2\up{2.0} & 53.3\up{2.9} & 47.5\up{1.1} \\
  & TACoT & 50.6\up{0.1} & 47.1\up{0.3} & 46.5\up{0.2} & 50.9\up{1.7} & 50.5\up{0.1} & 47.3\up{0.9} \\
\bottomrule
\end{tabular}
\caption{Comparison of mitigation strategies on KMMLU (accuracy in \%). Each accuracy column reports the mean over five perturbation intensity levels; colored annotations indicate gain/loss relative to Standard (\textcolor{blue}{blue}: improvement, \textcolor{red}{red}: degradation). Sub.: Jamo Substitution; Jong.: Jongseong Deletion; Rep.: Jamo Repetition; Space.: Space Deletion; Trans.: Jamo Transposition.}
\vspace{-2mm}
\label{tab:mitigation_results}
\end{table*}

\subsection{Baselines}
\label{sec:mitigation_baselines}

We compare TACoT against four inference strategies. 
\textbf{Standard} uses the default multiple-choice prompt and generates at most 8 tokens.
\textbf{Meta-Cognition} uses a system prompt that explicitly warns the model that the question may contain Korean typos, but still requires a direct single-letter answer.
\textbf{GEC} first applies KoGEC~\citep{kim-etal-2024-kogec} correction to the input and then uses the standard prompt. 
Although KoGEC can repair simple exposed-jamo errors, it is not designed for intra-syllabic jamo perturbations and may leave valid-form typos unchanged or introduce semantic alterations (Section~\ref{sec:correction_limits}); we nonetheless include it as a relevant correction-oriented baseline.
\textbf{CoT} applies the CoT prompt to every input with a maximum generation length of 1024 tokens. 
The exact system prompts for Standard, Meta-Cognition, and CoT inference are shown in Appendix~\ref{sec:appendix_prompt}.

\subsection{Results}
\label{sec:mitigation_results}

\begin{figure}[h]
    \centering
    \includegraphics[width=\columnwidth]{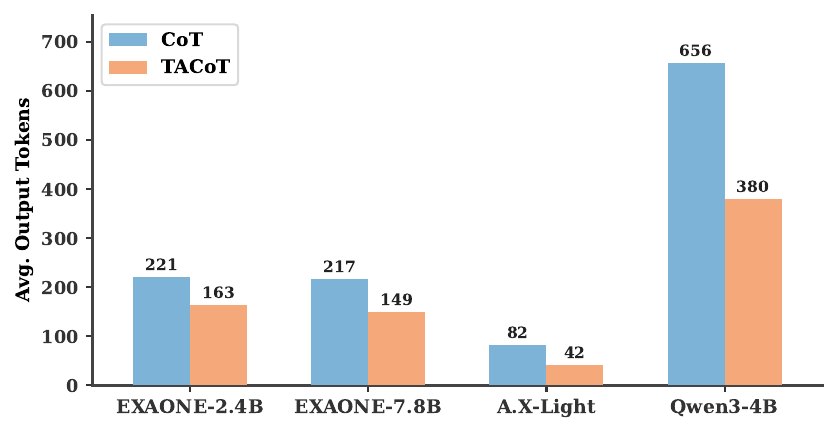}
    \caption{Average output tokens for CoT and TACoT across models. TACoT consistently reduces token cost while preserving most of the accuracy gains of CoT.}
    \label{fig:token_comparison}
\end{figure}

Table~\ref{tab:mitigation_results} compares the mitigation methods in terms of accuracy, and Figure~\ref{fig:token_comparison} summarizes the average output token cost of CoT and TACoT across models.
We report accuracy on clean inputs (\textit{Base}) and on each typo type, where typo accuracies are averaged over five perturbation intensity levels.

Overall, GEC-based pre-correction and Meta-Cognition prompting provide little to no reliable improvement over Standard inference.
In several cases, they even reduce accuracy.
For example, GEC lowers performance on Jamo Substitution for all four models, and Meta-Cognition prompting degrades Qwen3-4B across all typo types.
This suggests that Korean jamo typos are not reliably handled either by applying a general-purpose correction model before inference or by simply warning the model that typos may be present.

CoT is the most consistently effective mitigation strategy.
Across models, it improves accuracy not only on typo-corrupted inputs but also on clean inputs.
The gains are especially large for A.X-Light, where CoT improves the clean baseline from 49.5\% to 57.1\% and raises all typo-type accuracies by more than 5.9 percentage points.

TACoT provides a more efficient compromise.
By invoking CoT only when the hidden-state probe detects a likely typo, TACoT preserves most of the accuracy gains of CoT across all models and typo types.
As shown in Figure~\ref{fig:token_comparison}, this selective routing reduces average output length by approximately 37\% compared to CoT, ranging from 26\% for EXAONE-2.4B to 49\% for A.X-Light.

The remaining gap between TACoT and CoT is informative.
TACoT is strongest on jamo-level perturbations such as Jamo Repetition and Jamo Transposition, which also produce clear representational separation in Section~\ref{sec:hs_detection}.
In contrast, Space Deletion remains difficult for the routing mechanism: although CoT can still improve Space Deletion accuracy, this perturbation induces weak representational separation and is therefore less likely to trigger CoT routing.
These results show that the representational typo signal is not only diagnostic but also practically useful for routing, enabling cost-aware mitigation without sacrificing most of the accuracy gains that full CoT provides.

\begin{table*}[th]
\centering
\small
\setlength{\tabcolsep}{12pt}
\begin{tabular}{llcccccc}
\toprule
\textbf{Model} & \textbf{Method} & \textbf{Base} & \textbf{Sub.} & \textbf{Jong.} & \textbf{Rep.} & \textbf{Space.} & \textbf{Trans.} \\
\midrule
\multirow{4}{*}{EXAONE-2.4B}
  & Standard & 31.5 & 25.6 & 24.9 & 29.0 & 30.7 & 24.2 \\
  & Meta-Cognition & 30.2\dn{1.3} & 25.2\dn{0.4} & 25.0\up{0.1} & 28.4\dn{0.6} & 30.4\dn{0.3} & 23.4\dn{0.8} \\
  & CoT & 36.5\up{5.0} & 30.2\up{4.6} & 29.3\up{4.4} & 33.7\up{4.7} & 36.0\up{5.3} & 28.5\up{4.3} \\
  & TACoT & 31.9\up{0.4} & 29.4\up{3.8} & 28.2\up{3.3} & 33.2\up{4.2} & 33.0\up{2.3} & 28.3\up{4.1} \\
\midrule
\multirow{4}{*}{EXAONE-7.8B}
  & Standard & 37.7 & 32.5 & 31.8 & 35.8 & 37.2 & 30.6 \\
  & Meta-Cognition & 36.9\dn{0.8} & 32.3\dn{0.2} & 31.8 & 35.1\dn{0.7} & 37.0\dn{0.2} & 29.6\dn{1.0} \\
  & CoT & 45.4\up{7.7} & 39.1\up{6.6} & 38.3\up{6.5} & 43.0\up{7.2} & 44.6\up{7.4} & 36.8\up{6.2} \\
  & TACoT & 37.7 & 37.9\up{5.4} & 36.7\up{4.9} & 42.0\up{6.2} & 39.3\up{2.1} & 36.6\up{6.0} \\
\midrule
\multirow{4}{*}{A.X-Light}
  & Standard & 40.4 & 36.4 & 35.7 & 39.9 & 40.7 & 36.6 \\
  & Meta-Cognition & 40.5\up{0.1} & 36.4 & 35.9\up{0.2} & 39.7\dn{0.2} & 40.4\dn{0.3} & 36.9\up{0.3} \\
  & CoT & 48.0\up{7.6} & 43.1\up{6.7} & 42.1\up{6.4} & 47.0\up{7.1} & 47.8\up{7.1} & 43.3\up{6.7} \\
  & TACoT & 41.7\up{1.3} & 42.3\up{5.9} & 41.1\up{5.4} & 46.1\up{6.2} & 43.9\up{3.2} & 43.1\up{6.5} \\
\midrule
\multirow{4}{*}{Qwen3-4B}
  & Standard & 34.0 & 26.6 & 26.5 & 31.7 & 34.0 & 27.3 \\
  & Meta-Cognition & 18.0\dn{16.0} & 11.8\dn{14.8} & 12.6\dn{13.9} & 14.2\dn{17.5} & 16.1\dn{17.9} & 12.5\dn{14.8} \\
  & CoT & 42.2\up{8.2} & 35.4\up{8.8} & 35.1\up{8.6} & 40.8\up{9.1} & 42.7\up{8.7} & 36.1\up{8.8} \\
  & TACoT & 37.7\up{3.7} & 35.1\up{8.5} & 34.6\up{8.1} & 40.5\up{8.8} & 39.8\up{5.8} & 36.0\up{8.7} \\
\bottomrule
\end{tabular}
\caption{Accuracy (\%) on HRM8K, averaged over five intensity levels per typo type. Colored annotations indicate gain/loss relative to Standard (\textcolor{blue}{blue}: improvement, \textcolor{red}{red}: degradation). Sub.: Jamo Substitution; Jong.: Jongseong Deletion; Rep.: Jamo Repetition; Space.: Space Deletion; Trans.: Jamo Transposition. GEC is excluded, as KoGEC degenerates on mathematical inputs, producing empty or repetition-looped outputs even for clean questions.}
\label{tab:hrm8k_mitigation}
\end{table*}

\subsection{Robustness of TACoT}
\label{sec:generalization}

We examine the robustness of TACoT along two axes: whether it generalizes to a different domain and task format, and whether its gain reflects the value of the probe's routing choices rather than simply invoking CoT more often.

To test whether our findings hold beyond closed-form multiple-choice QA, we evaluate on HRM8K~\citep{ko2025understand}, a Korean mathematical reasoning benchmark requiring free-form solutions rather than a single-letter answer. 
See Appendix~\ref{sec:appendix_datasets} for dataset details. 
We apply the same five jamo-level perturbations to the question text.

Table~\ref{tab:hrm8k_mitigation} shows that the same trends observed on KMMLU carry over to this different domain and task: typos degrade accuracy relative to Base, CoT is the strongest mitigation, and TACoT recovers most of its gain. 

We also examine whether TACoT's gain comes from the probe's routing choices, or simply from invoking CoT more often. 
To isolate this, we compare TACoT against a rate-matched random router: the same number of inputs are routed to CoT per typo condition, chosen at random rather than by the probe. 
We restrict this comparison to the 7,311 questions all four models answer correctly when clean, so that CoT offers no benefit on clean inputs and any gain must come from recovering typo-corrupted ones.

As shown in Table~\ref{tab:rate_matched_gap}, the gap between TACoT and the random router is negligible at the mildest intensity level, where the perturbation is too weak for CoT to recover anything; from there it grows monotonically, reaching up to 2.3 points at the highest level. 
This is consistent with a valid routing signal: where direct inference fails most and the probe detects corruption most reliably, TACoT sends CoT precisely to the damaged inputs a random router misses.

\begin{table}[h]
\centering
\small
\setlength{\tabcolsep}{4pt}
\begin{tabular}{lccccc}
\toprule
\textbf{Model} & $l_1$ & $l_2$ & $l_3$ & $l_4$ & $l_5$ \\
\midrule
EXAONE-2.4B & +0.27 & +0.76 & +1.35 & +1.71 & +2.28 \\
EXAONE-7.8B & $-$0.09 & +0.08 & +0.60 & +1.12 & +1.67 \\
A.X-Light   & 0.00 & +0.20 & +0.70 & +1.06 & +1.63 \\
Qwen3-4B    & $-$0.03 & +0.19 & +0.51 & +0.60 & +0.77 \\
\bottomrule
\end{tabular}
\caption{Accuracy gap between TACoT and the random router, in points, on the base-correct pool, averaged over the four jamo-level typo types and five random seeds.}
\label{tab:rate_matched_gap}
\end{table}

\section{Conclusion}
This work systemically analyzes the impact of Korean typographical errors on LLMs. 
Unlike ordinary character-level edits, we show that Korean typos occur within the intra-syllabic jamo composition, consistently degrading model performance regardless of scale or language specialization. 
We demonstrate that these typos induce a distinct, identifiable shift in internal representations that is separate from ordinary answer incorrectness, allowing a simple linear probe to generalize across unseen error types. 
Leveraging this signal, we propose TACoT, a routing framework that achieves cost-aware mitigation by selectively invoking chain-of-thought inference. 
Overall, our findings establish Korean typo robustness as a crucial linguistic bottleneck and a valuable testbed for examining input corruption within LLM hidden states.

\section*{Limitations}
Our study has several limitations. 
Our perturbations are automatically generated from predefined typo types and intensity levels. 
Although they are grounded in Korean keyboard input and Korean jamo structure, they may not fully match the distribution of real user typos. 
Similarly, evaluation resources for Korean LLM robustness remain relatively limited compared with English, and broader coverage across domains, task formats, and naturally occurring typo distributions would strengthen future analyses.
On the modeling side, our probing and TACoT experiments rely on access to internal hidden states, which restricts their applicability to locally-served open-weight models; this is fundamentally incompatible with API-based LLM deployment, the most common real-world usage scenario. 
For this reason, our main probing and mitigation experiments are conducted on four small-to-mid-scale models (2.4B--7.8B), while the larger models in Appendix~\ref{sec:appendix_robustness} are evaluated only for accuracy degradation, without corresponding probe or TACoT validation.
Beyond model scale, TACoT is also a routing-based mitigation method: it detects typo-corrupted inputs and improves downstream accuracy by invoking more careful reasoning, but it does not directly correct or realign the typo-induced representational shift itself.
Developing methods that explicitly repair such representational distortions remains an important direction for future work.

\section*{Ethics Statement}

This work uses publicly available benchmark data and does not involve personal or sensitive information. 
The typo perturbations are synthetic and are designed to evaluate model robustness under realistic Korean input noise. 
While such perturbations could potentially be used to stress or evade deployed NLP systems, our goal is to characterize and mitigate this vulnerability. 
We report aggregate model behavior and propose a defensive routing method intended to improve robustness against typo-corrupted inputs.

\section*{Acknowledgments}

This work was supported by the Institute of Information \& Communications Technology Planning \& Evaluation (IITP) grant funded by the Korea government (MSIT) (No. RS-2021-II211341; Artificial Intelligence Graduate School Program at Chung-Ang University) and (No. RS-2026-25546026; Leading Generative AI Human Resources Development).
This work was also supported by the National Research Foundation of Korea (NRF) grant funded by the Korean government (MSIT) (RS-2026-25494299).
This work used datasets from `The Open AI Dataset Project (AI-Hub, S. Korea)'. All data information can be accessed through `AI-Hub (www.aihub.or.kr)'.

\bibliography{custom}

\newpage
\appendix

\section{Dataset Details}
\label{sec:appendix_datasets}

\paragraph{KMMLU.}
KMMLU~\citep{son2025kmmlu} is a Korean expert-level multiple-choice benchmark containing 35,030 four-choice questions across 45 subjects, including humanities, social sciences, and STEM.
Unlike Korean benchmarks translated from English, KMMLU is collected from original Korean examinations and therefore provides a natural evaluation setting for Korean language understanding.
We use KMMLU as the main evaluation benchmark throughout the paper: measuring accuracy under controlled typo perturbations (Section~\ref{sec:impact}), evaluating held-out typo detection on unseen questions (Section~\ref{sec:hs_detection}), and comparing mitigation methods (Section~\ref{sec:mitigation}).
KMMLU is never used for layer selection, probe training, or threshold tuning.

\paragraph{HAERAE-GK.}
For calibration, we use the General Knowledge subset of HAERAE Bench (HAERAE-GK)~\citep{son-etal-2024-hae}, a Korean knowledge benchmark whose source is disjoint from KMMLU.
HAERAE-GK contains 176 four-choice questions.
We apply the same five typo taxonomies and five perturbation intensity levels to this split, and use it for all fitting or selection steps: Fisher layer selection (Section~\ref{sec:hidden_signals}), probe training for held-out typo detection (Section~\ref{sec:hs_detection}), and probe training and threshold selection for TACoT (Section~\ref{sec:typo-aware_cot}).
This design keeps the KMMLU test questions unseen during calibration and prevents leakage from evaluation data.

\paragraph{HRM8K.}
To test generalization beyond KMMLU, we additionally evaluate on HRM8K~\citep{ko2025understand}, a Korean mathematical reasoning benchmark of 8,011 problems spanning multiple sources, including translated and native Korean math competitions.
Unlike KMMLU, HRM8K requires a free-form generated solution rather than a single-letter answer, and its subject matter (mathematical word problems) is disjoint from the encyclopedic and professional knowledge covered by KMMLU.
We apply the same five typo taxonomies and five perturbation intensity levels to the question text, and use HRM8K exclusively for evaluation (Section~\ref{sec:generalization}): the probe and TACoT routing threshold are still trained on HAERAE-GK, with no HRM8K questions used for fitting or selection.

\section{Perturbation Generation Details}
\label{sec:appendix_perturbation_algorithms}

We detail how each of the five perturbation types (Section~\ref{sec:korean_typographical_errors}) is generated, and release the full implementation so the benchmark can be regenerated exactly.

For a Hangul syllable $s$, $\textsc{Decompose}(s)=(c,v,f)$ returns its onset, nucleus, and (possibly empty) coda, with inverse $\textsc{Compose}$. 
The number of edits for type $t$ at level $\ell \in \{5,10,15,20,25\}\%$ is $k = \max(1, \lfloor n_t(x)\cdot\ell\cdot0.05\rfloor)$, where $n_t(x)$ is the number of spaces (T4) or complete Hangul syllables (otherwise). 
Levels are applied independently to the clean input.

T1, T3, and T5 share one structure: select $k$ eligible sites without replacement, decompose each, apply a type-specific edit, and recompose (Algorithm~\ref{alg:shared}).

\begin{algorithm}[h]
\caption{Shared structure for T1, T3, T5}
\label{alg:shared}
\begin{algorithmic}[1]
\Procedure{Perturb}{$x$, $k$, \textit{eligible}}
  \State $I \gets \textsc{SelectSites}(x, k, \textit{eligible})$
  \ForAll{$i \in I$}
    \State $(c, v, f) \gets \textsc{Decompose}(x_i)$
    \State $x_i \gets \textsc{Edit}(c, v, f)$
  \EndFor
  \State \Return $x$
\EndProcedure
\end{algorithmic}
\end{algorithm}

The type-specific edit differs as follows. 
T1 (Substitution) shuffles the component order and replaces the first component that has an adjacency-map entry with a uniformly sampled candidate. 
T3 (Repetition) picks one component at random and emits it twice: once inside the recomposed syllable, once as a standalone jamo appended after it. 
T5 (Transposition) swaps the onset and coda if a coda exists, or the onset and nucleus otherwise, producing a raw jamo sequence rather than a valid syllable; since some syllables are invariant under transposition, $2k$ sites are oversampled and the procedure stops once $k$ edits have actually changed the input.

T2 (Jongseong Deletion) and T4 (Space Deletion) do not follow this structure. 
T2 drops the coda of syllables that have one: $\textsc{Edit}(c,v,f) = \textsc{Compose}(c,v,\varepsilon)$ (\begin{CJK}{UTF8}{mj}한\end{CJK} $\rightarrow$ \begin{CJK}{UTF8}{mj}하\end{CJK}). 
T4 selects space characters instead of syllables and removes them, merging adjacent eojeol (\begin{CJK}{UTF8}{mj}한국 사회\end{CJK} $\rightarrow$ \begin{CJK}{UTF8}{mj}한국사회\end{CJK}).

Substitution candidates (T1) come from the standard Dubeolsik Korean keyboard layout: each key's horizontal and diagonal neighbors, keeping only the two nearest by key-center distance when more remain. Full adjacency maps are released with our implementation.

\section{Naturalness of the Perturbation Taxonomy}
\label{sec:appendix_taxonomy_validation}

To verify that our five perturbation types reflect real-world typing behavior rather than an artificial construction, we analyze the AI-Hub "High-Frequency Error Correction Data by Interface (Keyboard/Voice)"~\citep{aihub_typo_correction}, a corpus of real Korean keyboard and voice input errors containing 729,755 annotated error spans. 
We map it to our five perturbation types.

Table~\ref{tab:aihub_coverage} shows that our taxonomy accounts for 96.2\% of real errors, with only 3.8\% unclassified. 
Jamo-fragmenting errors (Repetition + Transposition) constitute 16.8\% of keyboard typos, confirming that structural jamo corruption is a genuine, frequent phenomenon rather than a synthetic artifact. 
The most frequent real jamo confusions (e.g., \begin{CJK}{UTF8}{mj}ㅣ\end{CJK}$\leftrightarrow$\begin{CJK}{UTF8}{mj}ㅏ\end{CJK} adjacent-key, \begin{CJK}{UTF8}{mj}ㅔ\end{CJK}$\leftrightarrow$\begin{CJK}{UTF8}{mj}ㅐ\end{CJK} near-homophone) also match the keyboard-layout assumptions underlying our substitution model.

\begin{table*}[h]
\centering
\small
\setlength{\tabcolsep}{6pt}
\begin{tabular}{lcccccc}
\toprule
\textbf{Error Source} & \textbf{Sub.} & \textbf{Jong.} & \textbf{Rep.} & \textbf{Space.} & \textbf{Trans.} & \textbf{Unclass.} \\
\midrule
All      & 34.4\% & 18.9\% & 9.5\%  & 50.1\% & 1.8\% & 3.8\% \\
Keyboard & 39.5\% & 19.6\% & 12.9\% & 43.5\% & 3.9\% & 0.9\% \\
\bottomrule
\end{tabular}
\caption{Coverage of real Korean input errors by our taxonomy, on 729,755 error spans from AI-Hub. \textit{All} includes both keyboard and voice input errors; \textit{Keyboard} restricts to the keyboard-only subset. Each error may carry multiple labels, so rows can exceed 100\%.}
\label{tab:aihub_coverage}
\end{table*}

\section{Prompt and Inference Details}
\label{sec:appendix_prompt}

Figures~\ref{fig:prompt_standard}, \ref{fig:prompt_typoaware}, and \ref{fig:prompt_cot} show the system prompts used in our experiments. The user message appends the question, answer choices formatted as \texttt{(A)}--\texttt{(E)}, and a Korean answer cue. GEC-based inference uses the standard prompt with the corrected question as input, applying the publicly released KoGEC model~\citep{kim-etal-2024-kogec} to typo-perturbed questions before standard inference.

All models are served via vLLM with greedy decoding at temperature~0; non-CoT generations use a maximum of 8 new tokens and CoT generations use a maximum of 1024 new tokens.
Hidden states are extracted using HuggingFace Library.
The logistic regression probe is implemented in scikit-learn with L2 regularization (\texttt{C=1.0}) on standardized features.

\tcbset{
  promptappendix/.style={
    colback=blue!3,
    colframe=blue!45!black,
    colbacktitle=blue!55!black
  }
}

\begin{figure}[h]
\centering
\begin{CJK}{UTF8}{mj}
\begin{promptbox}[promptappendix]{Standard Prompt}
당신은 한국어 시험 문제를 푸는 AI입니다.\\
\textit{(You are an AI that solves Korean exam questions.)}

\vspace{3pt}
주어진 선택지 중 정답의 알파벳(대문자 한 글자)만 출력하세요.\\
\textit{(Output only the letter of the correct answer in uppercase.)}
\end{promptbox}
\end{CJK}
\caption{Standard system prompt used for baseline inference and 
GEC-based inference.}
\label{fig:prompt_standard}
\end{figure}

\begin{figure}[h]
\centering
\begin{CJK}{UTF8}{mj}
\begin{promptbox}[promptappendix]{Meta-Cognition Prompt}
당신은 한국어 시험 문제를 푸는 AI입니다.\\
\textit{(You are an AI that solves Korean exam questions.)}

\vspace{3pt}
문제에 한국어 오타가 포함되어 있을 수 있습니다.\\
\textit{(The question may contain Korean typographical errors.)}

\vspace{3pt}
오타를 고려하여 문제의 의도를 파악한 뒤, 정답의 알파벳(대문자 한 글자)만 출력하세요.\\
\textit{(Considering the typos, infer the intended meaning and output only the correct answer letter in uppercase.)}
\end{promptbox}
\end{CJK}
\caption{Meta-Cognition system prompt used for meta-cognition inference.}
\label{fig:prompt_typoaware}
\end{figure}

\begin{figure}[h]
\centering
\begin{CJK}{UTF8}{mj}
\begin{promptbox}[promptappendix]{CoT Prompt}
당신은 한국어 시험 문제를 푸는 AI입니다.\\
\textit{(You are an AI that solves Korean exam questions.)}

\vspace{3pt}
문제에 오타가 포함되어 있을 수 있습니다.\\
\textit{(The question may contain typographical errors.)}

\vspace{3pt}
문제의 의도를 파악하기 위해 단계적으로 분석한 뒤, 마지막 줄에 '정답: X' 형식으로 알파벳 한 글자만 출력하세요.\\
\textit{(Analyze the question step by step to infer its intended meaning, then output only the answer letter in the format `Answer: X' on the last line.)}
\end{promptbox}
\end{CJK}
\caption{Chain-of-thought system prompt used for full-CoT and 
typo-aware mitigation inference.}
\label{fig:prompt_cot}
\end{figure}

\section{Additional Results for Typo Robustness}
\label{sec:appendix_robustness}

\begin{figure}[tb]
    \centering
    \includegraphics[width=\columnwidth]{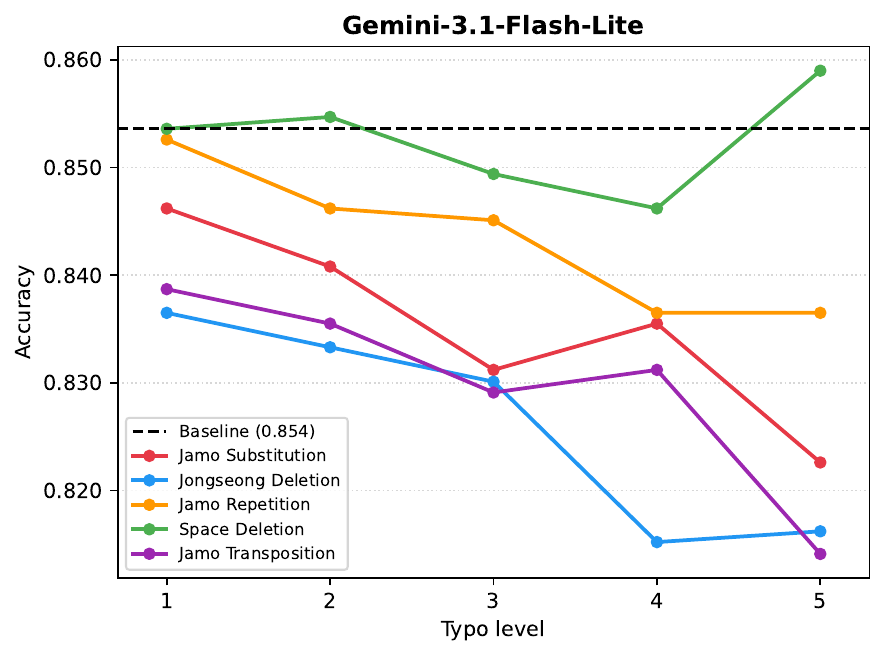}\\[6pt]
    \includegraphics[width=\columnwidth]{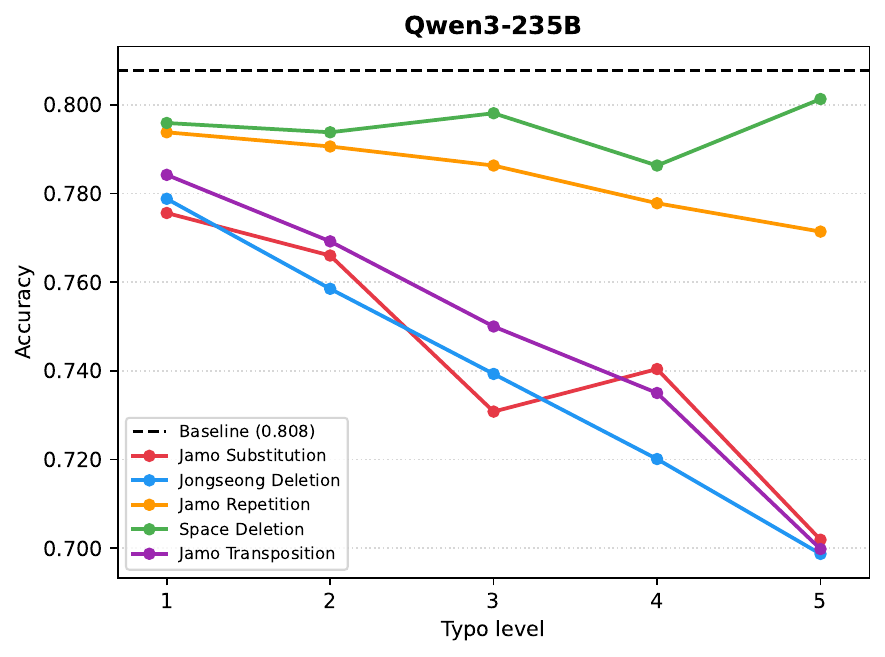}
    \caption{KMMLU accuracy under Korean typo perturbations for API-scale models. Dashed lines indicate clean baseline accuracy.}
    \label{fig:api_results}
\end{figure}

One might further expect larger or more capable models to be more robust to typographical noise. 
Figure~\ref{fig:api_results} evaluates two API-scale models, Gemini-3.1-Flash-Lite~\citep{google2026gemini31flashlite} and Qwen3-235B-A22B-2507~\citep{yang2025qwen3}, on the same perturbations. 
The degradation pattern persists: jamo-level perturbations consistently cause larger accuracy drops than Space Deletion across both models, suggesting that Korean typo vulnerability is not a limitation of model scale or training regime.

Beyond model scale, another possible source of variation is the perturbation procedure itself, since affected syllables are selected at random within each intensity level (Section~\ref{sec:typos_across_writing_systems}). We verify that this randomness does not drive our findings by regenerating the entire benchmark under two additional seeds and re-evaluating all four models; the corruption rate is fixed by the intensity, so only which syllables are affected varies across seeds.

Table~\ref{tab:seed_robustness} reports the mean accuracy over all 25 typo conditions (5 types $\times$ 5 levels) under three seeds. The mean moves by at most 0.07 points across seeds, and even the largest single-condition deviation is negligible: in the worst case across all models and conditions, Jongseong Deletion at level 3 on EXAONE-7.8B, accuracy ranges only from 43.26 to 43.83, a 0.57-point spread. No condition shows a meaningful seed effect.

\begin{table}[h]
\centering
\small
\setlength{\tabcolsep}{6pt}
\begin{tabular}{lcccc}
\toprule
\textbf{Model} & \textbf{Seed 1} & \textbf{Seed 2} & \textbf{Seed 3} & \textbf{Max $\Delta$} \\
\midrule
EXAONE-2.4B & 39.76 & 39.76 & 39.72 & 0.04 \\
EXAONE-7.8B & 44.11 & 44.05 & 44.08 & 0.06 \\
A.X-Light   & 47.17 & 47.24 & 47.22 & 0.07 \\
Qwen3-4B    & 47.79 & 47.79 & 47.77 & 0.02 \\
\bottomrule
\end{tabular}
\caption{Mean accuracy over all 25 typo conditions, under three random seeds for perturbation sampling.}
\label{tab:seed_robustness}
\end{table}

\section{Additional Representational Analyses}
\label{sec:appendix_hidden}

\subsection{Layer and Type Selection via Fisher Score}

\begin{figure*}[tb]
    \centering
    \includegraphics[width=\linewidth]{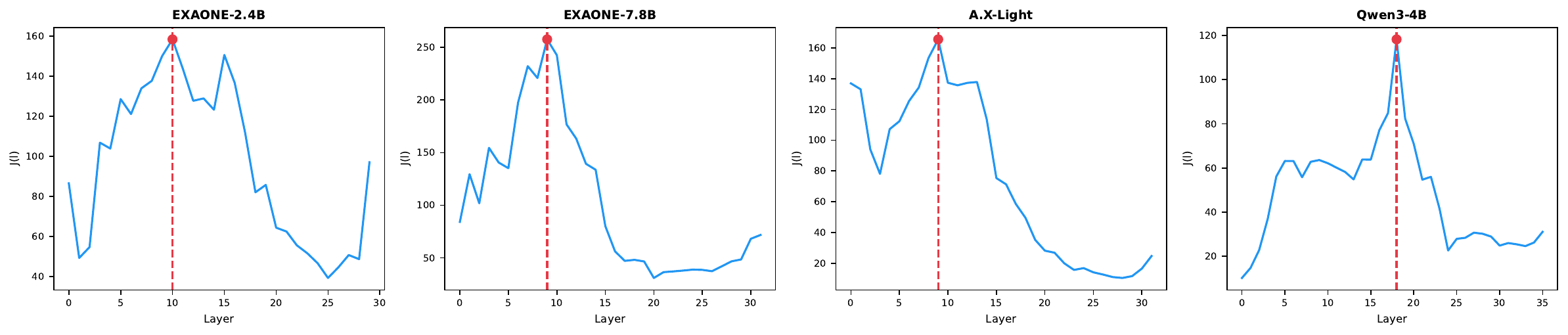}
    \caption{Layer-wise Fisher separation score $J(l)$ between clean and typo hidden states. Dashed vertical lines indicate the selected typo-sensitive layer for each model.}
    \label{fig:fisher_curve}
\end{figure*}

Figure~\ref{fig:fisher_curve} plots the Fisher separation score across transformer layers.
The curves show that typo sensitivity is not uniformly distributed across the model: each model has a localized region where clean and typo representations are more clearly separated.
We select the typo-sensitive layer from these peaks and use the selected layer for the subsequent direction and probe analyses.
The selected layers are layer 10 for EXAONE-2.4B, layer 9 for EXAONE-7.8B, layer 9 for A.X-3.1-Light, and layer 18 for Qwen3-4B.

Table~\ref{tab:fisher_per_type} reports the Fisher separation score at the selected layer for each typo type, averaged over perturbation intensities. 
Space Deletion consistently yields the lowest scores across all models, with an average of 0.009 compared to 0.052--0.164 for jamo-level types, confirming that it induces qualitatively weaker representational shifts.
Among jamo-level types, Jamo Transposition produces the largest separation, while Jongseong Deletion produces the smallest --- a pattern that mirrors their relative probe detection difficulty in Table~\ref{tab:hs_typo_detection}.

\begin{table}[t]
\centering
\small
\setlength{\tabcolsep}{3.5pt}
\begin{tabular}{lccccc}
\toprule
\textbf{Model} & \textbf{Sub.} & \textbf{Jong.} & \textbf{Rep.} & \textbf{Space.} & \textbf{Trans.} \\
\midrule
EXAONE-2.4B & 0.062 & 0.061 & 0.103 & 0.010 & 0.220 \\
EXAONE-7.8B & 0.112 & 0.093 & 0.115 & 0.021 & 0.230 \\
A.X-Light   & 0.034 & 0.035 & 0.039 & 0.004 & 0.093 \\
Qwen3-4B    & 0.023 & 0.020 & 0.055 & 0.001 & 0.112 \\
\midrule
Avg.        & 0.058 & 0.052 & 0.078 & 0.009 & 0.164 \\
\bottomrule
\end{tabular}
\caption{Fisher separation score $J$ at the Fisher-selected layer, averaged over perturbation intensities. Abbreviations denote Jamo Substitution (Sub.), Jongseong Deletion (Jong.), Jamo Repetition (Rep.), Space Deletion (Space.), and Jamo Transposition (Trans.).}
\label{tab:fisher_per_type}
\end{table}

\subsection{Detection Robustness Across Intensity}

Section~\ref{sec:hs_detection} reports held-out detection AUROC pooled over all five perturbation intensities (5\%--25\%), showing strong detection across all four models. 
Table~\ref{tab:auroc_by_intensity} breaks this same held-out probe down by intensity level, averaged over the four held-out jamo-level typo types, and the same conclusion holds at a finer granularity: even at the weakest level $l_1$, AUROC is 0.79--0.85, far above chance, and rises monotonically with intensity.
If the probe were merely detecting semantic incoherence rather than the typo itself, its performance should drop sharply at low intensities where sentence coherence is largely preserved; instead, detection is already strong at $l_1$, indicating that the probe responds to the perturbation itself.

\begin{table}[h]
\centering
\small
\setlength{\tabcolsep}{4pt}
\begin{tabular}{lccccc}
\toprule
\textbf{Model} & $l_1$ & $l_2$ & $l_3$ & $l_4$ & $l_5$ \\
\midrule
EXAONE-2.4B & 0.789 & 0.883 & 0.942 & 0.972 & 0.984 \\
EXAONE-7.8B & 0.846 & 0.921 & 0.966 & 0.984 & 0.991 \\
A.X-Light   & 0.826 & 0.924 & 0.975 & 0.991 & 0.997 \\
Qwen3-4B    & 0.798 & 0.880 & 0.940 & 0.972 & 0.985 \\
\bottomrule
\end{tabular}
\caption{Held-out detection AUROC by perturbation intensity, averaged over the four held-out jamo-level typo types.}
\label{tab:auroc_by_intensity}
\end{table}

\subsection{Isolating the Typo Signal from Answer Correctness}

\begin{figure*}[t]
\centering
\includegraphics[width=0.48\linewidth]{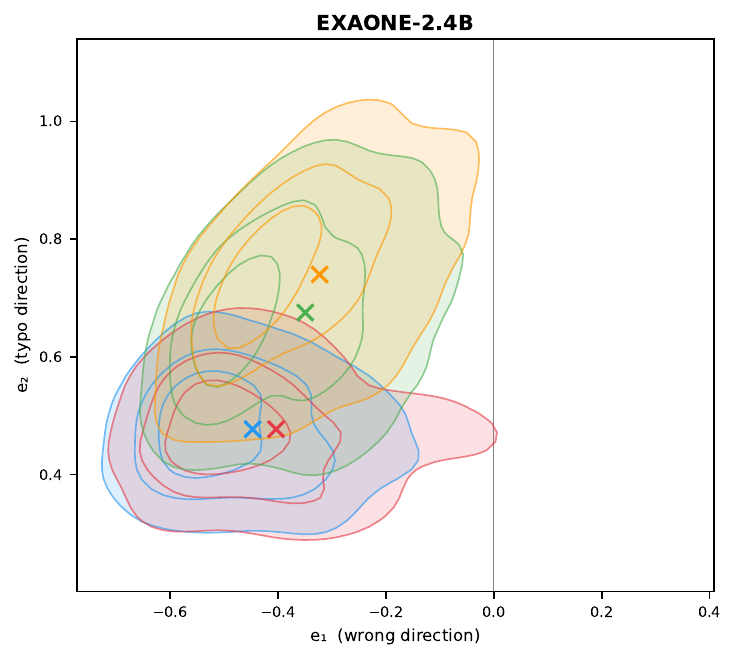}
\hfill
\includegraphics[width=0.48\linewidth]{figures/direction_proj_LGAI-EXAONE_EXAONE-3.5-7.8B-Instruct.pdf}

\vspace{4pt}

\includegraphics[width=0.48\linewidth]{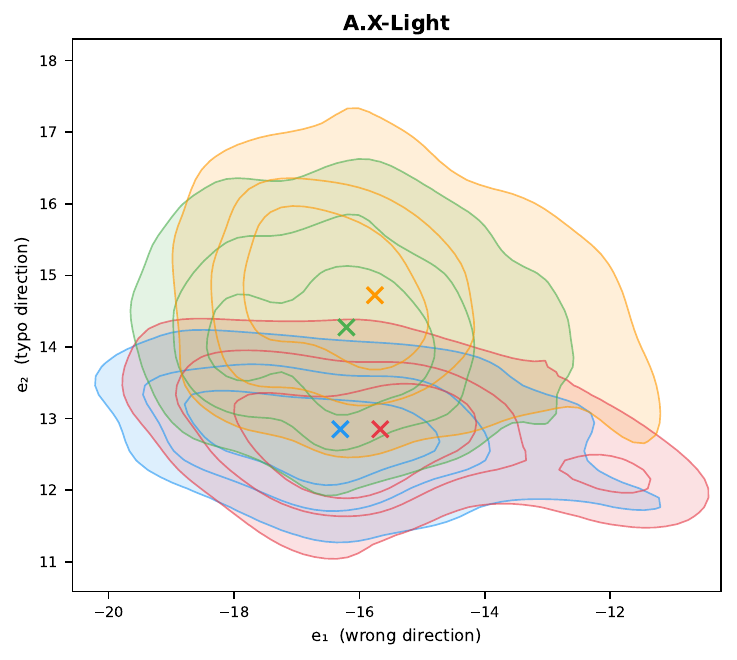}
\hfill
\includegraphics[width=0.48\linewidth]{figures/direction_proj_Qwen_Qwen3-4B-Instruct-2507.pdf}

\vspace{3pt}
\small
\textcolor{blue}{\rule{1em}{0.8pt}} Base-Correct \quad
\textcolor{red}{\rule{1em}{0.8pt}} Base-Wrong \quad
\textcolor{green!60!black}{\rule{1em}{0.8pt}} Typo-Correct \quad
\textcolor{orange}{\rule{1em}{0.8pt}} Typo-Wrong

\caption{Full direction projections of hidden states at the Fisher-selected typo-sensitive layer for all evaluated models. The horizontal axis corresponds to the clean incorrectness direction, while the vertical axis captures the typo direction orthogonal to it.}
\label{fig:direction_projection_all}
\end{figure*}

Figure~\ref{fig:direction_projection_all} visualizes hidden states after projecting them onto two directions: the clean incorrectness direction and the typo direction orthogonal to it.
If typo-induced errors were merely ordinary answer failures, typo examples would collapse mainly along the clean incorrectness axis.
Instead, both typo-correct and typo-wrong examples consistently shift along the orthogonal typo direction.
This supports the main-text claim that Korean typos induce a distinct representational displacement rather than only increasing the likelihood of an incorrect answer.

\section{Probe Training Details}
\label{sec:appendix_probe}

For held-out typo detection (Section~\ref{sec:hs_detection}), we train a logistic regression probe on the last-token hidden state at the Fisher-selected layer, using clean inputs and typo-perturbed inputs from HAERAE-GK as binary labels. 
The training set is balanced by randomly sampling typo examples to match the number of clean inputs, drawing only from the non-held-out typo types.

For TACoT (Section~\ref{sec:typo-aware_cot}), we split the balanced HAERAE-GK calibration data into training and validation subsets with an 80/20 split. 
The probe is trained with standardized hidden states using logistic regression. 
The routing threshold $\theta$ is selected by maximizing Youden's $J$ statistic~\citep{youden1950index}, $J = \mathrm{TPR} - \mathrm{FPR}$, on the validation split. 
No KMMLU examples are used for either probe training or threshold selection.

\section{Case Study}
\label{sec:appendix_case}

We provide representative recovery cases where Standard inference predicts an
incorrect answer, while CoT inference recovers the correct answer. The examples
cover four jamo-level typo types: Jamo Substitution, Jongseong Deletion, Jamo
Repetition, and Jamo Transposition. For readers unfamiliar with Korean, we
briefly describe the role of the CoT trajectory before each figure.

Figure~\ref{fig:case_ax_sub} shows a Jamo Substitution case. The corrupted
question asks for the output voltage of an amplifier, but several Korean words
are misspelled. Standard inference selects the wrong option, whereas CoT
recovers the intended voltage-gain calculation and obtains the correct answer.

Figure~\ref{fig:case_exaone78_jong} shows a Jongseong Deletion case. The input
contains valid-looking but corrupted Korean forms, so the typo is not visually
obvious as an invalid string. CoT nevertheless extracts the numerical quantities
and performs the unit conversion needed to answer the question.

Figure~\ref{fig:case_exaone24_rep} shows a Jamo Repetition case. The repeated
jamo exposes an abnormal Korean character sequence inside the word for
``screw.'' CoT explicitly identifies the corrupted form and uses the surrounding
context to infer that the question refers to a micrometer.

Figure~\ref{fig:case_qwen_trans} shows a Jamo Transposition case. The typo
breaks multiple syllable blocks and exposes raw jamo throughout the question.
CoT first reconstructs the likely clean question, then reasons over the answer
choices to identify the only output device.

\tcbset{
  casestudybox/.style={
    enhanced,
    colback=gray!3,
    colframe=gray!55!black,
    colbacktitle=gray!65!black,
    fontupper=\footnotesize,
    boxrule=0.5pt,
    top=4pt,
    bottom=4pt,
    left=5pt,
    right=5pt
  }
}

\begin{figure*}[p]
\centering
\begin{minipage}{0.98\textwidth}
\begin{CJK}{UTF8}{mj}
\begin{promptbox}[casestudybox]{A.X-Light: Jamo Substitution}
\textbf{Typo input}\\[2pt]
40[db]의 잔압이득를 가짅 증폭기에 10[mV]의 전압을 입력에 가하면 출력전압은 몇 [V] 인가?

\vspace{2pt}
\textit{[An amplifier with a voltage gain of 40[dB] has an input of 10[mV]. What is the output voltage in [V]?]}

\vspace{4pt}
\textbf{Choices}\\[2pt]
(A) 0.1[V] \quad (B) 1[V] \quad (C) 10[V] \quad (D) 100[V]

\vspace{4pt}
\textbf{Standard output}\\[2pt]
C

\vspace{4pt}
\textbf{CoT trajectory}\\[2pt]
이 문제를 풀기 위해 증폭기의 전압 이득(전압 이득 = 출력 전압 / 입력 전압)을 이용하겠습니다.

주어진 정보:
\begin{itemize}[leftmargin=1.3em,itemsep=0pt,topsep=1pt]
\item 잔압(전압 이득) = 40 dB
\item 입력 전압 = 10 mV
\end{itemize}

먼저, dB(데시벨)로 표현된 전압 이득을 선형 값으로 변환해야 합니다.
전압 이득(선형) = $10^{(\mathrm{dB}/20)}$

따라서:
전압 이득(선형) = $10^{(40/20)} = 10^2 = 100$

이제 출력 전압을 계산할 수 있습니다:
출력 전압 = 입력 전압 $\times$ 전압 이득(선형)

출력 전압 = 10 mV $\times$ 100 = 1000 mV = 1 V

\vspace{2pt}
\textit{[To solve this, I use the voltage gain formula (gain = output / input). Given: voltage gain = 40 dB, input = 10 mV. Converting to linear: $10^{(40/20)} = 100$. Output = 10 mV $\times$ 100 = 1000 mV = 1 V.]}

\vspace{4pt}
\textbf{CoT output}\\[2pt]
B
\end{promptbox}
\end{CJK}
\end{minipage}
\caption{Case study for A.X-Light under Jamo Substitution. Standard inference predicts an incorrect answer, while CoT inference recovers the correct answer.}
\label{fig:case_ax_sub}
\end{figure*}

\begin{figure*}[p]
\centering
\begin{minipage}{0.98\textwidth}
\begin{CJK}{UTF8}{mj}
\begin{promptbox}[casestudybox]{EXAONE-7.8B: Jongseong Deletion}
\textbf{Typo input}\\[2pt]
노선의 전체의 기이가 2km 인 다각측량을 실시하였더니, 폐합비가 1/5000 이어다.
폐합오차는 며 cm 인가?

\vspace{2pt}
\textit{[A traverse survey on a route of total length 2km yielded a closure ratio of 1/5000. What is the closure error in cm?]}

\vspace{4pt}
\textbf{Choices}\\[2pt]
(A) 0.04 cm \quad
(B) 0.4 cm \quad
(C) 4 cm \quad
(D) 40 cm

\vspace{4pt}
\textbf{Standard output}\\[2pt]
B

\vspace{4pt}
\textbf{CoT trajectory}\\[2pt]
문제를 분석해보면, 주어진 정보는 다음과 같습니다:

\begin{itemize}[leftmargin=1.3em,itemsep=0pt,topsep=1pt]
\item 노선의 전체 길이: 2km
\item 폐합비: 1/5000
\end{itemize}

폐합비는 측량에서 전체 길이에 대한 허용 오차 비율을 나타냅니다.
따라서 폐합오차를 계산하는 공식은 다음과 같습니다:

$\text{폐합오차} = \text{전체 길이} \times \text{폐합비}$

단위를 맞추기 위해 전체 길이를 미터로 변환합니다:
$2\text{km} = 2000\text{m}$

이제 계산을 진행합니다:
$\text{폐합오차} = 2000\text{m} \times \frac{1}{5000} = 0.4\text{m}$

미터를 센티미터로 변환하면:
$0.4\text{m} = 40\text{cm}$

따라서 정답은:

\vspace{2pt}
\textit{[Analyzing the problem, the given information is as follows: total route length = 2km, closure ratio = 1/5000. The closure ratio represents the allowable error relative to the total length in surveying. Therefore, closure error = total length $\times$ closure ratio. Converting to meters: 2km = 2000m. Closure error = 2000m $\times$ 1/5000 = 0.4m. Converting to centimeters: 0.4m = 40cm. Therefore, the answer is:]}

\vspace{4pt}
\textbf{CoT output}\\[2pt]
D
\end{promptbox}
\end{CJK}
\end{minipage}
\caption{Case study for EXAONE-7.8B under Jongseong Deletion. Standard inference predicts an incorrect answer, while CoT inference recovers the correct answer.}
\label{fig:case_exaone78_jong}
\end{figure*}

\begin{figure*}[p]
\centering
\begin{minipage}{0.98\textwidth}
\begin{CJK}{UTF8}{mj}
\begin{promptbox}[casestudybox]{EXAONE-2.4B: Jamo Repetition}
\textbf{Typo input}\\[2pt]
정확한 피치의 나ㅅ사를 이용하여 실제ㅇ의 길이를 측정하는 측정용 기기는 어느 것이ㅣㄴ가?

\vspace{2pt}
\textit{[Which measuring instrument uses a screw with a precise pitch to measure actual length?]}

\vspace{4pt}
\textbf{Choices}\\[2pt]
(A) 버니어 캘리퍼스 \quad
(B) 마이크로 미터 \quad
(C) 높이 게이지 \quad
(D) 다이얼 게이지

\vspace{2pt}
\textit{[(A) Vernier calipers \quad
(B) Micrometer \quad
(C) Height gauge \quad
(D) Dial gauge]}

\vspace{4pt}
\textbf{Standard output}\\[2pt]
A

\vspace{4pt}
\textbf{CoT trajectory}\\[2pt]
문제에서 "나ㅅ사"가 오타로 보입니다.
문맥상 정확한 피치의 나사를 이용한 길이 측정에 초점을 맞추고 있는 것 같으나,
주어진 옵션들은 주로 직접적인 길이 측정 도구들입니다.
나사의 피치를 통해 간접적으로 길이를 측정하는 개념이 포함되어 있을 수 있지만,
주어진 선택지들은 모두 직접적인 길이 측정 도구들입니다.

주어진 옵션들 중에서 가장 정확한 길이 측정에 특화된 도구는:

\vspace{2pt}
\textit{[``나ㅅ사'' appears to be a typo for ``나사'' (screw). The question focuses on length measurement using a precise-pitch screw. Among the options, the instrument most specialized for precise length measurement via screw pitch is the micrometer.]}

\vspace{4pt}
\textbf{CoT output}\\[2pt]
B
\end{promptbox}
\end{CJK}
\end{minipage}
\caption{Case study for EXAONE-2.4B under Jamo Repetition. Standard inference predicts an incorrect answer, while CoT inference recovers the correct answer.}
\label{fig:case_exaone24_rep}
\end{figure*}

\begin{figure*}[p]
\centering
\begin{minipage}{0.98\textwidth}
\begin{CJK}{UTF8}{mj}
\begin{promptbox}[casestudybox]{Qwen3-4B: Jamo Transposition}
\textbf{Typo input}\\[2pt]
컴퓨터ㅏㄱ 이해할 수 ㅇㅆㅣㄴㄴㅡ 형태로 데ㅣㅇ터ㄹㄹㅡ 변환하여 입력시키는 ㅈㅇㅏ치가 아닌 것은?

\vspace{2pt}
\textit{[Which of the following is NOT a device that converts data into a form understandable by a computer?]}

\vspace{4pt}
\textbf{Choices}\\[2pt]
(A) 키보드 \quad
(B) 광마우스 \quad
(C) 디지타이저 \quad
(D) XY플로터

\vspace{2pt}
\textit{[(A) Keyboard \quad
(B) Optical mouse \quad
(C) Digitizer \quad
(D) XY plotter]}

\vspace{4pt}
\textbf{Standard output}\\[2pt]
C

\vspace{4pt}
\textbf{CoT trajectory}\\[2pt]
문제를 단계적으로 분석해 보겠습니다.

\textbf{문제 내용 요약:}

"컴퓨터ㅏㄱ 이해할 수 ㅇㅆㅣㄴㄴㅡ 형태로 데ㅣㅇ터ㄹㄹㅡ 변환하여 입력시키는 ㅈㅇㅏ치가 아닌 것은?"

이 문장은 명백히 오타가 많고, 자연스러운 한국어가 아닙니다.
이를 정리해보면, 다음과 같은 의미로 해석할 수 있습니다:

"컴퓨터가 이해할 수 있는 형태로 데이터를 변환하여 입력시키는 장치가 아닌 것은?"

즉, 컴퓨터에 데이터를 입력하는 장치 중에서 가장 적절하지 않은 것을 고르는 문제입니다.

각 선택지를 살펴보겠습니다:

\textbf{(A) 키보드}: 키를 누르면 컴퓨터가 이해할 수 있는 코드로 변환하여 입력합니다. 적절한 입력 장치입니다.
\textbf{(B) 광마우스}: 움직임을 X, Y 좌표로 변환하여 입력합니다. 적절한 입력 장치입니다.
\textbf{(C) 디지타이저}: 인간의 입력을 디지털 신호로 변환합니다. 적절한 입력 장치입니다.
\textbf{(D) XY플로터}: 데이터를 기반으로 물리적 형태로 출력합니다. 입력 장치가 아니라 \textbf{출력 장치}입니다.

따라서, "컴퓨터가 이해할 수 있는 형태로 데이터를 변환하여 입력시키는 장치가 아닌 것"은: (D) XY플로터

\vspace{2pt}
\textit{[The question contains many typos. Reconstructed: ``Which is NOT a device that converts data into a form a computer can understand?'' (A) Keyboard: input device. (B) Optical mouse: input device. (C) Digitizer: input device. (D) XY plotter: outputs data physically --- an \textbf{output device}, not an input device.]}

\vspace{4pt}
\textbf{CoT output}\\[2pt]
D
\end{promptbox}
\end{CJK}
\end{minipage}
\caption{Case study for Qwen3-4B under Jamo Transposition. Standard inference predicts an incorrect answer, while CoT inference recovers the correct answer.}
\label{fig:case_qwen_trans}
\end{figure*}

\section{Use of AI Assistants}
\label{sec:appendix_ai}

We authored this manuscript ourselves; AI assistance was used solely for refinement of style, clarity, and grammar, and not for ideation or content generation.


\end{document}